\documentclass[times, preprint, 10pt]{elsarticle}
\usepackage{setspace}
\usepackage{natbib}
\usepackage{graphicx}
\usepackage{hyperref}
\usepackage{url}
\usepackage{float}
\usepackage{booktabs}
\usepackage{amsfonts}
\usepackage{nicefrac}
\usepackage{microtype}
\usepackage{xcolor}
\usepackage{subcaption}
\usepackage{array}    
\usepackage{multirow} 

\renewcommand{\textcolor}[2]{#2}
\DeclareRobustCommand{\rev}[1]{#1}

\usepackage{amssymb}
\usepackage{amsmath}

\begin{document}
\singlespacing

\begin{frontmatter}



\title{Wisteria: A Unified Multi-Scale Feature Learning Framework for DNA Language Model}

%


\author[a,b,c]{Weihua Wang\corref{cor1}}
\author[a]{Haoji Li}
\author[a,b,c]{Feilong Bao}
\author[d]{Lei Yang}
\author[a,b,c]{Guanglai Gao}

\cortext[cor1]{Corresponding author. Email: wangwh@imu.edu.cn}

\affiliation[a]{organization={College of Computer Science, Inner Mongolia University},%
            city={Hohhot}, 
            state={Inner Mongolia},
            postcode={010021}, 
            country={China}}
\affiliation[b]{organization={National and Local Joint Engineering Research Center of Intelligent Information Processing Technology for Mongolian},%
            city={Hohhot}, 
            state={Inner Mongolia},
            postcode={010021}, 
            country={China}}
\affiliation[c]{organization={Inner Mongolia Key Laboratory of Multilingual Artificial Intelligence Technology},%
            city={Hohhot}, 
            state={Inner Mongolia},
            postcode={010021}, 
            country={China}}
\affiliation[d]{organization={School of Life Sciences, Inner Mongolia University},%
            city={Hohhot}, 
            state={Inner Mongolia},
            postcode={010020}, 
            country={China}}

\begin{abstract}
DNA language model aims to decipher the regulatory grammar and semantic of genomes by capturing long range dependencies in DNA sequences. 
Existing methods emphasize long range token interactions but often ignore the interplay between local motifs and global dependencies. 
In this paper, we propose Wisteria, a genomic language model that integrates multi scale feature learning within a unified framework for DNA sequence.
Specifically, Wisteria augments the Mamba based architecture with gated dilated convolutions to capture local motifs and regulatory patterns, while gated multilayer perceptrons refine global dependencies. 
We further introduce a Fourier based attention mechanism to support frequency domain modeling, periodic extension and length generalization.
Across four experimental settings with both short and long range dependencies, Wisteria demonstrates strong performance on downstream benchmarks against competitive DNA language model baselines.
These results indicate that Wisteria effectively unifies local and global dependency modeling for multi scale genomic sequence analysis.
\end{abstract}



\begin{keyword}

DNA language model \sep State Space model \sep Gated convolutions \sep Fourier Position Embedding 


\end{keyword}

\end{frontmatter}

\section{Introduction}

Genomic language models \rev{aim} to decipher the regulatory grammar of genomes \rev{by modeling how regulatory elements interact and how these interactions influence gene expression.}
Because genomic regulation is highly complex, many studies have focused on specific downstream tasks, such as identifying regulatory elements like promoters or enhancers~\cite{zhang2022ipro}, classifying mutations as deleterious or benign~\cite{pei2021deepfun}, and predicting transcription factor binding sites~\cite{wang2022towards}, splice sites~\cite{jaganathan2019predicting,zeng2022predicting}, gene expression~\cite{avsec2021effective}, and chromatin accessibility~\cite{trieu2020deepmilo}. 
\rev{Recent work has further explored task specific biological sequence analysis, including transcription factor binding site localization from DNA sequences with an interpretable dual channel encoder-decoder structure~\cite{ding2025deeputf} and interpretable identification of RNA $N^6$-methyladenosine modification sites by integrating sequence and structural information~\cite{li2025bijective}.}
These tasks collectively highlight the promise of genomic language modeling, but they also expose a common challenge: different genomic predictions depend on different sequence scales.
For instance, tasks like splice site recognition and motif discovery primarily depend on local contiguous DNA sequences or adjacent regulatory elements, 
typically spanning tens to hundreds of nucleotide base pairs (bp). 
In contrast, tasks such as enhancer identification or variant effect prediction require modeling long range dependencies spanning thousands to even megabases of base pairs. 
To fully decipher the regulatory grammar of genomes, genomic language models must simultaneously capture fine-grained sequence features and genome-wide regulatory contexts.

Recent genomic models built \rev{on} the long range Mamba architecture~\cite{gu2023mamba, schiff2024caduceus} exploit its selective state space mechanism to model long contexts efficiently, substantially improving the representation of long range dependencies.
However, state space model (SSM) inspired models tend to focus on continuous and global context, making them less effective at emphasizing localized features.
These local features, such as promoter regions or splice junctions, are essential for transcriptional regulation~\cite{schoenfelder2019long}.

In addition to local and global dependencies, DNA sequences are known to exhibit intrinsic periodic \rev{signals. For example,} dinucleotide or A/T-tract oscillations with a periodicity of approximately 10--11 bp are closely associated with nucleosome positioning, chromatin organization, and gene expression regulation~\cite{mrazek2010comparative,frenkel2011nucleosome}.
Therefore, it is essential for genomic language models to capture such frequency domain patterns to reflect, at least to some degree, hidden positional regularities.
Although Rotary Position Embedding (RoPE)~\cite{su2024roformer} can capture certain periodic signals, it suffers from spectral damage during periodic extension on long-context learning~\cite{huafourier}, which weakens the model’s ability to effectively capture long range genomic dependencies.

To address these challenges, we propose Wisteria
~\footnote{We name our model Wisteria, inspired by the natural growth pattern of the wisteria vine. Just as wisteria intertwines multiple branches into a cohesive and adaptive structure, 
our model integrates convolutional, BiMamba, and Fourier based attention mechanisms into a unified architecture. }, 
a unified genomic modeling framework that integrates local and global modeling through three key components.
First, we integrate gated dilated convolutions to enhance local motif recognition and enable multi scale receptive fields. 
Second, we incorporate gated MLPs to refine the global features produced by the Mamba block, ensuring that critical regulatory relationships are preserved through selective gating. 
Third, we introduce a Fourier based attention mechanism that integrates Fourier Position Embedding (FoPE)~\cite{huafourier} to enhance frequency domain modeling.
Unlike RoPE, FoPE expresses positional information as a Fourier series with multiple harmonic frequencies. 
This design alleviates the spectral damage caused by nonlinear transformations and improves the model’s ability to capture both periodic and long range genomic dependencies, enabling stronger length generalization and more stable global contextual alignment.
By integrating gated convolutions, gated MLPs, and Fourier based attention into bidirectional Mamba blocks, Wisteria achieves a unified representation that bridges local motif extraction and long range regulatory modeling. 
Extensive evaluations demonstrate strong overall performance across both short and long sequence genomic benchmarks, highlighting its ability to balance local precision with global understanding.

In summary, our contributions are as follows:
\begin{itemize}
\item We propose Wisteria, a unified genomic modeling architecture that integrates gated convolutions, gated MLPs, and a Fourier based attention mechanism into the Mamba framework, enabling effective learning of both local and global dependencies in DNA sequences.  
\item We introduce a novel incorporation of Fourier Position Embedding (FoPE) into genomic attention modeling, which enhances periodic signal representation and improves the model’s spectral smoothness and length generalization.
To the best of our knowledge, this is the first work to integrate FoPE into a genomic language model, enabling frequency domain reasoning within genomic sequence understanding.
\item We validate Wisteria across a range of genomic benchmarks encompassing both short and long sequence tasks. The results demonstrate that Wisteria achieves state of the art performance on the majority of evaluated tasks.
\end{itemize}

\section{Related work}
Genomic language models often follow a pretraining and fine-tuning paradigm~\cite{shu2026survey} to identify the intricate genomic grammar and distant regulatory interactions.
During the pretraining phase, they learn base co-occurrence patterns, conserved motif structures,
and distant regulatory dependencies from massive unlabeled genomic sequences via
self-supervised objectives such as masked language modeling (MLM) or next-token prediction.
The resulting pretrained models are then adapted to diverse downstream tasks through supervised fine-tuning, including variant effect prediction, regulatory element annotation, and gene expression modeling.
Existing genomic language models can be broadly categorized by their underlying architectures into
convolutional neural network models, Transformer based models, and state space models.

Convolutional neural network (CNN)-based models excel at capturing local dependencies and motifs in genomic sequences through the application of filters across input data. 
\rev{For example, DeepSEA~\cite{zhou2015predicting} demonstrated that convolutional architectures can predict chromatin features directly from sequence, 
Basset~\cite{kelley2016basset} showed strong performance on cell specific chromatin accessibility prediction, 
and DeepDBP-CNN~\cite{khan2020deepdbp} further applied CNN based modeling to DNA-binding protein prediction.}
GPN~\cite{benegas2023dna} performs effectively in predicting whole-genome variant effects in Arabidopsis thaliana with expanded CNN layers.
These models have demonstrated success in predicting DNA-protein binding sites, regulatory elements, and transcription factor binding sites (TFBS).  
CNN models' inherent local inductive bias enables superior data efficiency and strong performance on limited training data, making them ideal for small scale datasets. 
However, while techniques like dilation or downsampling allow handling of longer sequences, CNNs are inherently limited in modeling long range dependencies, 
especially in ultra-long DNA sequences exceeding 100 kilobases. 
This underscores the necessity for complementary methods to integrate local and global interactions in genomic data.

Transformer based models, such as DNABERT~\cite{ji2021dnabert}, {DNABERT-2}~\cite{zhou2024dnabert} and the Nucleotide Transformer~\cite{dalla2025nucleotide}, 
utilize attention mechanisms to enable each token to attend to all positions in the input sequence simultaneously. 
These models focus on relevant sequence parts and have advanced the understanding of regulatory mechanisms underlying gene expression~\cite{avsec2021effective,linder2025predicting}.
In addition, Enformer~\cite{avsec2021effective} integrates convolutional layers with attention mechanisms to capture both short and long range regulatory interactions, 
enhancing gene expression prediction.
Nevertheless, Transformer models lack inductive biases toward local interactions~\cite{zaheer2020big,su2024roformer}, reducing their data efficiency for local motifs like TFBS. 
Their quadratic complexity also renders them impractical for extremely long genomic sequences~\cite{dai2019transformer}.
For example, the input length of the NT-v2 model was limited to 12 kb~\cite{dalla2025nucleotide}.
Although attention variants, like BigBird~\cite{zaheer2020big} and GENA-LM~\cite{fishman2023gena}, reduce complexity to sub-quadratic levels and extend inputs to 36 kb, 
this often compromises performance by sacrificing full pairwise attention.

Position embeddings are critical for Transformer models because self-attention is permutation invariant. 
Early absolute position encodings provide fixed signals, but they risk overfitting to specific sequence lengths, as in sinusoidal functions~\cite{vaswani2017attention}.
Relative approaches like ALiBi~\cite{presstrain} add distance based biases, while Rotary Position Embedding (RoPE)~\cite{su2024roformer} uses complex rotations for relative dependencies.
Yet, RoPE's single frequency per dimension restricts spectral expressiveness, leading to degradation after transformations and poor length generalization.
FoPE~\cite{huafourier} addresses these by modeling each dimension as a multi-frequency Fourier series, enhancing spectral representation and reducing leakage during propagation. 
This allows simultaneous encoding of short and long range dependencies. 
Through multi-frequency modeling, FoPE achieved superior robustness and length generalization, scaling to 16 times the training length in needle-in-a-haystack tests.
FoPE outperformed RoPE and ALiBi across model sizes.

As an efficient alternative to Transformers, state space models (SSMs) provide near-linear scaling with sequence length. 
For example, HyenaDNA~\cite{nguyen2023hyenadna} based on the Hyena Hierarchy~\cite{poli2023hyena}, supported contexts up to 1 million nucleotides with a compact model size compared with Transformers. 
EVO~\cite{nguyen2024sequence} combined Hyena and Transformer architectures, pretraining on 8 kb sequences and fine-tuning on 131 kb during context extension. 
Caduceus~\cite{schiff2024caduceus}, built on Mamba~\cite{gu2023mamba}, trained on 131 kb sequences with reverse-complement augmentation and reported stronger downstream performance 
than HyenaDNA and several Transformer based models.
However, SSM based models are limited in fine-grained local feature extraction.

In light of these complementary strengths and limitations, we pursue a unified approach that learns semantic dependencies in DNA sequences from both local and global perspectives.
\section{Method}
\label{Method}

\subsection{Framework Overview}
\label{sec:Model}
\begin{figure}[H]
    \centering
    \includegraphics[width=1\linewidth]{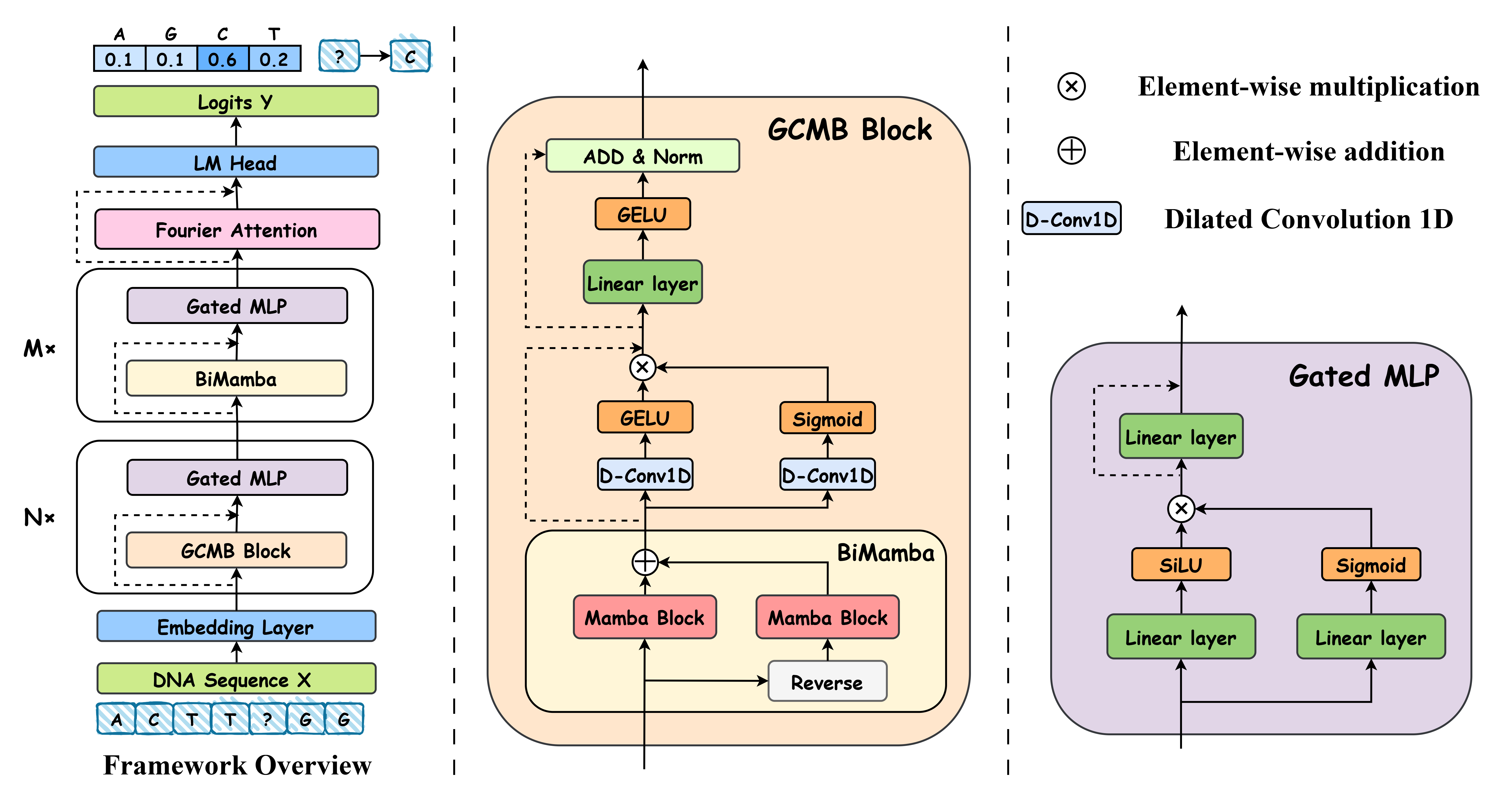}
    \caption{\textbf{The framework of the proposed Wisteria.} 
    The architecture is consist of Gated Convolution--BiMamba (GCMB) modules, gated MLP modules, and a final Fourier based attention layer. }
    \label{fig:framework}
\end{figure}

Our framework adopts a bottom-up \rev{design in which} shallow layers focus on fine-grained convolutional patterns, middle layers emphasize sequence-level dynamics via bidirectional state modeling, and the top layer refines global contextual interactions in the frequency domain.
This architecture is intended to model both short range and long range dependencies within genomic sequences in a unified manner.
The input DNA sequence initialized by one-hot encoding is first transformed into a dense embedding space through a learnable embedding layer.
Subsequently, feature extraction proceeds through a stack of Gated Convolution--BiMamba (GCMB) modules and gated MLP modules, each equipped with residual connections to support gradient stability and feature reuse.
Finally, a Fourier based attention mechanism in the last layer aggregates multi scale representations in the frequency domain, yielding position aware and semantically enriched sequence embeddings.
Figure~\ref{fig:framework} presents the overall framework of Wisteria. 

\subsection{Model Architecture}

Formally, the proposed architecture comprises four major components:  
the embedding layer, the Gated Convolution--BiMamba (GCMB) module, the gated MLP module, and the Fourier based attention mechanism.  
Each component is described below.

\paragraph{Embedding Layer}
Given a DNA sequence $\mathbf{S} = [s_1, s_2, \dots, s_L]$, where $s_i \in \{A, T, C, G\}$, 
we map it into dense vector representations:
\begin{equation}
\mathbf{X} = \text{Embed}(\mathbf{S}), \quad \mathbf{X} \in \mathbb{R}^{L \times D}.
\end{equation}
Here, $\mathbf{X}$ denotes the token-level embedding matrix, where $L$ is the sequence length and $D$ is the embedding dimension.  
Unlike previous architectures, no positional encoding is added at this stage; spatial dependencies are instead learned implicitly through subsequent sequence modeling modules.

\paragraph{GCMB Module}
The Gated Convolution--BiMamba (GCMB) module is responsible for capturing both multi scale local structures and long range sequential dependencies.  
Given input $\mathbf{X} \in \mathbb{R}^{L \times D}$, the BiMamba block first encodes bidirectional dependencies using bidirectional Mamba streams:
\begin{equation}
\begin{aligned}
\overrightarrow{\mathbf{H}} &= \text{Mamba}_{\text{fwd}}(\mathbf{X}), \\
\overleftarrow{\mathbf{H}} &= \text{Mamba}_{\text{bwd}}(\text{Flip}(\mathbf{X})), \\
\mathbf{H} &= \frac{1}{2} \left(\overrightarrow{\mathbf{H}} + \text{Flip}(\overleftarrow{\mathbf{H}})\right),
\end{aligned}
\end{equation}
Here, $\overrightarrow{\mathbf{H}}$ and $\overleftarrow{\mathbf{H}}$ denote the forward and backward hidden representations, respectively, $\text{Mamba}_{\text{fwd}}(\cdot)$ and $\text{Mamba}_{\text{bwd}}(\cdot)$ are the two directional Mamba streams, and $\mathbf{H} \in \mathbb{R}^{L \times D}$ is their averaged bidirectional representation. The operator $\text{Flip}(\cdot)$ reverses the sequence order to model backward context.  
This bidirectional design allows the model to capture dependencies from both upstream and downstream nucleotides.

Two dilated depthwise convolutions then extract and gate local contextual information:
\begin{equation}
\begin{aligned}
\mathbf{h} &= \text{GeLU}\!\left(\text{convolution}_A(\mathbf{H})\right), \\
\mathbf{g} &= \sigma\!\left(\text{convolution}_B(\mathbf{H})\right),
\end{aligned}
\end{equation}
Here, $\mathbf{h} \in \mathbb{R}^{L \times D}$ is the activated local feature branch, $\mathbf{g} \in \mathbb{R}^{L \times D}$ is the gate branch, $\text{convolution}_A(\cdot)$ and $\text{convolution}_B(\cdot)$ denote two depthwise convolutions with different dilation rates, $\text{GeLU}(\cdot)$ is the Gaussian error linear unit, and $\sigma(\cdot)$ is the sigmoid function. These two convolutions use different dilation rates to model diverse receptive fields.  
The gated output is computed as:
\begin{equation}
\mathbf{Y} = \text{LayerNorm}\!\left(\text{MLP}\!\left(\mathbf{H} + \mathbf{h} \odot \mathbf{g}\right)\right), 
\quad \mathbf{Y} \in \mathbb{R}^{L \times D},
\end{equation}
where $\mathbf{Y}$ is the output feature matrix of the GCMB block, $\odot$ denotes element-wise multiplication, and $\text{MLP}(\cdot)$ and $\text{LayerNorm}(\cdot)$ denote the feed-forward projection and layer normalization, respectively.  
Note that MLP is applied before normalization to stabilize the gating effect.
In the shallow layers, the dilation rates follow an expansion schedule with a dilation factor of $n$, namely $[1, 1, n, n^2, n^3]$, allowing the model to detect both short motifs (e.g., 5--10 bp TATA boxes) and broader functional elements (10--200 bp).

\paragraph{Gated MLP Module}
Following GCMB, the gated MLP module performs nonlinear feature fusion and refinement:
\begin{equation}
[\mathbf{U}, \mathbf{G}] = \text{Split}\!\left(\mathbf{W}_1 \mathbf{Y} + \mathbf{b}_1\right), 
\quad \mathbf{U}, \mathbf{G} \in \mathbb{R}^{L \times h},
\end{equation}
Here, $\mathbf{W}_1$ and $\mathbf{b}_1$ are the learnable weight matrix and bias of the input projection, $\text{Split}(\cdot)$ divides the projected features into two equal branches, and $h$ is the expanded hidden dimension.  
Feature activation and gating are jointly applied as:
\begin{equation}
\tilde{\mathbf{U}} = \text{SiLU}(\mathbf{U}) \odot \sigma(\mathbf{G}),
\end{equation}
where $\tilde{\mathbf{U}} \in \mathbb{R}^{L \times h}$ is the gated hidden representation, $\text{SiLU}(\cdot)$ denotes the sigmoid linear unit, and $\sigma(\cdot)$ is the sigmoid function.  
The output is then computed with a residual connection but without normalization:
\begin{equation}
\mathbf{Z} = \mathbf{Y} + \mathbf{W}_2 \tilde{\mathbf{U}} + \mathbf{b}_2, 
\quad \mathbf{Z} \in \mathbb{R}^{L \times D}.
\end{equation}
Here, $\mathbf{W}_2$ and $\mathbf{b}_2$ are the learnable parameters of the output projection, and $\mathbf{Z}$ denotes the refined representation produced by the gated MLP module.

\paragraph{Fourier based Attention}
In the final layer, Wisteria adopts a Fourier based attention mechanism incorporating \textit{Fourier Position Embedding (FoPE)}. 
FoPE includes a frequency gating mechanism to suppress undertrained low-frequency components below a cutoff, preserving long wave information and spectral balance. 
It represents each embedding dimension as a Fourier series of multiple frequencies:
\begin{equation}
f(\omega_m) = 
\begin{cases}
1, & \omega_m < \omega_l, \\
e^{i\omega_m n} + \displaystyle\sum_{\omega} a_{\omega} e^{i\omega n}, & \omega_m \geq \omega_l,
\end{cases}
\end{equation}
Here, $f(\omega_m)$ denotes the Fourier positional response at frequency $\omega_m$, $\omega_m$ is the base frequency associated with one embedding component, $\omega_l = 2\pi / N$ is the low-frequency cutoff determined by the sequence length $N$, $n$ denotes the token position index, $i$ is the imaginary unit, and $a_\omega$ are learnable harmonic coefficients for additional frequency components.  
The attention function is defined as:
\begin{equation}
\text{Attn}(\mathbf{q}, \mathbf{k}) = 
\text{softmax}\!\left( \frac{(\mathbf{q} f(\omega)) (\mathbf{k} f(\omega))^\top}{\sqrt{D}} \right),
\end{equation}
where $\mathbf{q}$ and $\mathbf{k}$ denote the query and key representations, respectively, $f(\omega)$ applies the Fourier positional modulation to them, and $\sqrt{D}$ is the standard scaling factor based on the feature dimension.  
This formulation enhances periodic and long range interactions while maintaining spectral smoothness and positional awareness.

\paragraph{Summary}
In summary, Wisteria establishes a hierarchical modeling pipeline:  
an embedding layer for nucleotide representation,  
GCMB modules for multi scale convolutional and bidirectional sequence modeling,  
gated MLPs for nonlinear fusion and residual refinement, and  
a Fourier based attention mechanism for frequency domain contextual enhancement.  
This integrated design enables Wisteria to effectively unify local and global genomic dependencies, achieving robust generalization across sequence lengths and structural patterns.




\section{Experiments}
\label{sec:Experiment}
To evaluate the performance and generalization ability of Wisteria, we conduct a series of experiments covering pretraining, downstream tasks, and more fine-grained analyses.
We first pretrain the model on the human reference genome (Sec.~\ref{sec:pretraining}). 
We further evaluate it on a wide range of downstream benchmarks, including Genomic Benchmarks, Nucleotide Transformer tasks, variant effect prediction, and BEND (Sec.~\ref{sec:downstream}). These experiments assess both supervised and zero-shot capabilities.
\rev{We then conduct ablation studies on the Nucleotide Transformer tasks (Sec.~\ref{sec:ablation}).}
We also analyze the influence of positional encodings on length generalization, the contribution of \rev{the gated convolution branch to} local feature modeling, and the throughput and memory efficiency of different architectural configurations (Sec.~\ref{sec:analysis}).
\rev{Because the benchmarks differ in their task settings, the exact comparison set is not identical across all tables. 
In each case, we report the stronger baselines that are most compatible with the corresponding benchmark setting.}
Together, these experiments provide a holistic view of Wisteria's performance across diverse genomic contexts and sequence length scales.
All experiments are conducted on H20 GPUs with 96GB memory\rev{, with the exact GPU configuration depending on the model setting.}

\subsection{Masked Language Modeling (MLM) Pretraining}
\label{sec:pretraining}
To consistent with HyenaDNA and Caduceus, we also pretrained our model on the human reference genome hg38 (GRCh38). 
In this experiment, the data source is identical to HyenaDNA, and the data splitting method is consistent with Caduceus. 
The training set contains 34,021 genomic segments, each extended to a maximum length of 1,048,576 bases ($2^{20}$).
This training set covers the entire genome and has approximately 35 billion nucleotide tokens in total.

By default, our architecture employs a 12-layer design. 
The first five layers incorporate dilated gated convolutions alongside BiMamba blocks to form GCMB modules for multi scale local feature learning.  
The remaining layers rely on BiMamba and gated MLP blocks for deeper contextual modeling.  
In the final layer, the standard self-attention is replaced by a Fourier based attention mechanism that explicitly integrates frequency information via Fourier Position Embedding (FoPE).  
This hierarchical design enables Wisteria to efficiently bridge the gap between localized motif detection and global dependency modeling across entire DNA sequences.

DNA sequences were tokenized at the character level for two reasons. 
First, k-mer tokenization is highly sensitive to small shifts such as single-base insertions or deletions, which can produce entirely different tokenizations~\cite{zhou2024dnabert}. 
Second, variant effect prediction requires single nucleotide resolution, which demands fine-grained tokenization.

For the masking strategy, we adopted the BERT-style scheme~\cite{devlin2019bert}, masking 15\% of tokens. 
Of these, 80\% were replaced with the special [MASK] token, 10\% with random vocabulary tokens, and 10\% left unchanged.

To stabilize training across varying sequence lengths, we fixed the number of tokens per batch.
Specifically, sequences of length 1,024 used a batch size of 1,024, while sequences of length 131,072 ($2^{17}$) used a batch size of 8.
This scheme is identical to that adopted in Caduceus \rev{ and maintains an approximately constant token budget per batch, i.e., $\text{global batch size} \times \text{sequence length} \approx 1{,}048{,}576$ tokens.}

For the remaining optimization settings, \rev{we used a learning rate of $8 \times 10^{-3}$ for the 64-dim 1024 bp configuration and $1 \times 10^{-3}$ for the other two pretrained configurations.
Training used AdamW~\cite{loshchilovdecoupled} with a warmup cosine decay schedule, where $\beta_1$ and $\beta_2$ were set to 0.9 and 0.95, respectively. 
Key pretraining hyperparameters are summarized in Table~\ref{tab:pretraining_hyperparams}.}

\begin{table}[h]
\centering
\caption{Pretraining hyperparameters. Values used during pretraining are reported.}
\label{tab:pretraining_hyperparams}
\begin{tabular}{lc}
\toprule
\textbf{Hyperparameter} & \textbf{Value} \\
\midrule
Weight Decay & 0.1 \\
Dropout & 0.0 \\
Optimizer & AdamW \\
Optimizer Momentum & \textcolor{blue}{$\beta_1=0.9$, $\beta_2=0.95$} \\
Learning Rate Scheduler & \textcolor{blue}{Warmup Cosine Decay} \\
Precision & AMP \\
Attention Heads & 16 \\  
Kernel Size & 9 \\    
Dilation Base & 3 \\   
\bottomrule
\end{tabular}
\end{table}

\subsection{Downstream Tasks}
\label{sec:downstream}

\rev{To assess the generalization capability of Wisteria across diverse genomic tasks, we evaluate the model on four downstream benchmarks covering short and long sequence classification, variant effect prediction, and regulatory element annotation.}
The configurations used in the four downstream experiments are summarized in Table~\ref{tab:wisteria_experiments}. 
In terms of training time,
the 64-dim 1024 bp configuration was pretrained on one H20 GPU in about 30 minutes. 
The 768-dim 1024 bp configuration was pretrained on four H20 GPUs in about 2 days and 3 hours, 
and the 768-dim 131,072 bp configuration was pretrained on four H20 GPUs in about 4 days and 10 hours.

\begin{table}[htbp]
\centering
\caption{\rev{Wisteria configurations used in different downstream tasks.}}
\label{tab:wisteria_experiments}
\small
\resizebox{\textwidth}{!}{
\begin{tabular}{lccccc}
\toprule
\textcolor{blue}{\textbf{Task / Experiment}} & \textcolor{blue}{\textbf{Dim.}} & \textcolor{blue}{\textbf{Len.(bp)}} & \textcolor{blue}{\textbf{Learning Rate}} & \textcolor{blue}{\textbf{Steps}} & \textcolor{blue}{\textbf{Size}} \\
\midrule
\textcolor{blue}{Genomic Benchmarks} & \textcolor{blue}{64} & \textcolor{blue}{1024} & \textcolor{blue}{$8\times10^{-3}$} & \textcolor{blue}{10,000} & \textcolor{blue}{550K} \\
\textcolor{blue}{\shortstack[l]{Nucleotide Transformer tasks}} & \textcolor{blue}{768} & \textcolor{blue}{1024} & \textcolor{blue}{$1\times10^{-3}$} & \textcolor{blue}{50,000} & \textcolor{blue}{100M} \\
\textcolor{blue}{BEND} & \textcolor{blue}{768} & \textcolor{blue}{131,072} & \textcolor{blue}{$1\times10^{-3}$} & \textcolor{blue}{50,000} & \textcolor{blue}{100M} \\
\textcolor{blue}{\shortstack[l]{Variant effect prediction (VEP)}} & \textcolor{blue}{768} & \textcolor{blue}{131,072} & \textcolor{blue}{$1\times10^{-3}$} & \textcolor{blue}{50,000} & \textcolor{blue}{100M} \\
\bottomrule
\end{tabular}
}
\end{table}

\subsubsection{Genomic Benchmarks}
\label{sec:genomics}
To evaluate the ability of Wisteria to capture regulatory patterns in short genomic sequences, we conducted experiments on the Genomic Benchmarks datasets~\cite{grevsova2023genomic}.
The Genomic Benchmarks comprise eight regulatory element classification datasets with sequence lengths ranging from 200 to 2000 bp.
We employed 5-fold cross-validation. 
In each fold, the training set was further divided into a 90/10 split for training and validation, and early stopping on the validation split was used to prevent overfitting.
Fine-tuning was performed for up to 10 epochs, with fewer epochs in cases of early stopping. 
To reduce the effect of randomness, all experiments were repeated with five different random seeds, and we report both mean performance and the deviations.
For downstream classification, all models shared the same prediction head: the final hidden states were pooled across the sequence dimension, followed by a linear transformation to produce logits, which were then used for loss computation and prediction.

We compared Wisteria with three baseline models.
The \textbf{CNN} baseline was introduced by~\cite{grevsova2023genomic}.
It was trained from scratch with a learning rate of \(1 \times 10^{-3}\) and a batch size of \(64\).
Its architecture consists of an embedding layer followed by three one-dimensional convolutional layers with 16, 8, and 4 channels, and a fully connected classifier.
The \textbf{HyenaDNA} model~\cite{nguyen2023hyenadna} employs a non-Transformer architecture based on Hyena operators,
which combine long convolutions with implicit gating to achieve sub-quadratic complexity.
In this experiment, we reproduced the HyenaDNA with a hidden dimension of 128, supporting sequences up to 1,024 bp with single-nucleotide resolution.
It is fine-tuned with a learning rate of \(6 \times 10^{-4}\) and a batch size of \(256\).
Pretrained weights were obtained from the public checkpoint.~\footnote{\url{https://huggingface.co/LongSafari/hyenadna-tiny-1k-seqlen}}
The \textbf{Caduceus} model~\cite{schiff2024caduceus} is a BiMamba architecture.
In this experiment, we used the Caduceus-Ph configuration.
It contains four MambaDNA layers with a hidden dimension of 118 and uses base pair tokenization.
During fine-tuning, they used a learning rate of \(1 \times 10^{-3}\) and a batch size of \(128\).

To keep the parameter at the same order of magnitude with the above baselines,
our model used a 64 dimensional embedding and was pretrained on 1024bp sequences.
During fine-tuning phase, it used a learning rate of \(1 \times 10^{-3}\) and a batch size of 128.
Fine-tuning across all models was performed by AdamW, with the above mentioned model specific learning rates and batch sizes.

\begin{table}[htbp]
\centering
\caption{\rev{\textbf{Genomic Benchmarks results.} Top-1 accuracy (\(\uparrow\)) is reported across 5-fold cross-validation. 
Best values per task are bolded, second-best values are underlined, and \(\pm\) indicates the standard deviation across 5 random seeds
.}}
\label{tab:ft_comparison}
\resizebox{\textwidth}{!}{ 
\begin{tabular}{lcccc}
\hline
\textbf{Task} & \textbf{CNN} & \textbf{HyenaDNA} & \textbf{Caduceus-Ph} & \textbf{Wisteria} \\
\textbf{(Parameter Count)} & \textbf{(264K)} & \textbf{(436K)} & \textbf{(470K)} & \textbf{(550K)} \\
\hline
Mouse Enhancers & 0.730 $\pm$0.032 & \underline{0.779} $\pm$0.013 & 0.754 $\pm$0.074 & \textbf{0.795} $\pm$0.008 \\
Coding vs. Intergenomic & 0.892 $\pm$0.008 & 0.904 $\pm$0.008 & \underline{0.915 }$\pm$0.003 & \textbf{0.935} $\pm$0.002 \\
Human vs. Worm & 0.942 $\pm$0.002 & 0.961 $\pm$0.002 & \textbf{0.973} $\pm$0.001 & \underline{0.971} $\pm$0.001 \\
Human Enhancers Cohn & 0.702 $\pm$0.021 & 0.718 $\pm$0.008 & \textbf{0.747} $\pm$0.004 & \underline{0.745} $\pm$0.006 \\
Human Enhancer Ensembl & 0.744 $\pm$0.122 & 0.832 $\pm$0.006 & \underline{0.893 }$\pm$0.008 & \textbf{0.898} $\pm$0.005 \\
Human Regulatory & 0.872 $\pm$0.005 & 0.862 $\pm$0.004 & \underline{0.873} $\pm$0.011 & \textbf{0.887} $\pm$0.003 \\
Human OCR Ensembl & 0.698 $\pm$0.013 & 0.744 $\pm$0.019 & \underline{0.828} $\pm$0.006 & \textbf{0.829} $\pm$0.005 \\
Human Non-TATA Promoters & 0.861 $\pm$0.009 & 0.887 $\pm$0.005 & \underline{0.946} $\pm$0.007 & \textbf{0.949} $\pm$0.004 \\
\hline
\end{tabular}
}
\end{table}

Table~\ref{tab:ft_comparison} summarizes the Genomic Benchmarks results. Top-1 accuracy measures the proportion of correctly classified sequences among all predictions, with values closer to 1.0 indicating stronger classification performance.
Wisteria achieves the best performance on 6 of the 8 datasets, indicating strong modeling ability on short sequence classification tasks.
For example, the Coding vs Intergenomic task involves sequences of only 200 bp, while the Human Regulatory task operates on 401 bp sequences.

\subsubsection{Nucleotide Transformer Tasks}
\label{sec:nucleotide}
To evaluate the generalization capability of Wisteria across diverse genomic tasks, we conducted experiments on the Nucleotide Transformer benchmark~\cite{dalla2025nucleotide}. 
This benchmark integrates 18 tasks covering histone mark prediction, promoter identification, splice site recognition, and enhancer related prediction. 
These tasks cover three core bioinformatics categories: histone modification prediction, regulatory element annotation, and splice site identification. 

We compared Wisteria with four baseline models.
The \textbf{HyenaDNA} and \textbf{Caduceus-Ph} for this benchmark use larger configurations than those in the Genomic Benchmarks experiment.
Specifically, HyenaDNA used a 1.6M parameter model with 4 layers and a hidden dimension of 256.
Caduceus-Ph used a 1.9M parameter configuration with 4 layers and a hidden dimension of 256,
fine-tuned with a learning rate of \(1 \times 10^{-3}\) and task specific batch sizes of 128 or 512.
The \textbf{Nucleotide Transformer v2 (NT-v2)}~\cite{dalla2025nucleotide} employs a Transformer encoder architecture with a 12 kb context window.
It used 6-mer tokenization and was pretrained on 3,202 human genomes from the 1000 Genomes Project plus 850 diverse species using masked language modeling.
The \textbf{DNABERT-2}~\cite{zhou2024dnabert} is a Transformer based genomic language model that uses Byte Pair Encoding (BPE) tokenization.
It incorporated attention with linear biases for position encoding to generalize length beyond the pretrained sequence length.
The model was pretrained on multi-species genomic data.

For this experiment, we use the 768-dim configuration pretrained on 1024 bp sequences.
In our model, fine-tuning was performed using a learning rate of $1 \times \rev{10^{-3}}$ and a batch size of 256 for up to 10 epochs, with fewer epochs in cases of early stopping.

The datasets were taken from the public release~\footnote{\url{https://huggingface.co/spaces/InstaDeepAI/nucleotide_transformer_benchmark}}.
We follow the same protocol with baselines, including a 90/10 train/validation split within each fold.
The evaluation metrics follow the original benchmark: MCC (Matthews Correlation Coefficient) is used for histone mark and enhancer prediction tasks,
F1-score for promoter region and splice-site identification, and accuracy for the ``Splice All'' task.
MCC measures the quality of binary classifications taking into account true and false positives and negatives, with values ranging from $-1$ to $+1$, where $+1$ represents perfect prediction, $0$ indicates random guessing, and $-1$ indicates total disagreement.
The F1-score is the harmonic mean of precision and recall, ranging from 0 to 1, with higher values indicating better balance between correctly identifying positive cases and minimizing false positives and false negatives. 
Results are reported over 10 runs with different random seeds. 

\begin{table}[htbp]
\centering
\caption{\rev{\textbf{Nucleotide Transformer benchmark results.} MCC (\(\uparrow\)) is reported for histone mark and enhancer prediction tasks,
 F1-score (\(\uparrow\)) for promoter and splice-site identification tasks, and accuracy (\(\uparrow\)) for ``Splice All.'' 
 Best values per task are bolded, second-best values are underlined, and \(\pm\) indicates the variation across ten random seeds.}}
\label{tab:nt_benchmark}
\resizebox{\textwidth}{!}{
\begin{tabular}{l c c c c c} 
\hline 
\textbf{Task} & \textbf{NT-v2} & \textbf{HyenaDNA} & \textbf{DNABERT-2} & \textbf{Caduceus-Ph} & \textbf{Wisteria} \\  
\hline

\textbf{\textit{Histone}} & & & & & \\ \hline
H3 & 78.17 $\pm$2.54 & 78.14 $\pm$1.70 & 79.31 $\pm$0.68 & \underline{80.48} $\pm$1.04 & \textbf{84.47} $\pm$0.72 \\ 
H3K4me1 & 51.64 $\pm$1.12 & 44.52 $\pm$2.59 & 48.34 $\pm$4.63 & \underline{52.83} $\pm$0.96 & \textbf{59.39} $\pm$0.89 \\ 
H3K4me2 & 37.24 $\pm$2.25 & 42.68 $\pm$2.66 & 43.02 $\pm$2.92 & \underline{49.88} $\pm$2.65 & \textbf{55.36} $\pm$1.83 \\ 
H3K4me3 & 50.30 $\pm$1.77 & 50.41 $\pm$3.15 & 45.43 $\pm$3.33 & \underline{56.72 }$\pm$2.58 & \textbf{64.27} $\pm$0.85 \\ 
H3K9ac & 61.05 $\pm$1.40 & 58.50 $\pm$1.75 & 60.04 $\pm$1.27 & \underline{63.27 }$\pm$2.29 & \textbf{67.65} $\pm$1.55 \\ 
H3K14ac & 57.22 $\pm$2.19 & 56.71 $\pm$2.40 & 54.49 $\pm$4.99 & \underline{60.84} $\pm$2.94 & \textbf{70.16} $\pm$1.90 \\ 
H3K36me3 & 60.50 $\pm$1.75 & 59.92 $\pm$1.06 & 57.58 $\pm$2.38 & \underline{61.12} $\pm$1.44 & \textbf{71.55} $\pm$1.02 \\ 
H3K79me3 & 65.78 $\pm$2.34 & 66.25 $\pm$3.65 & 64.38 $\pm$0.48 & \underline{67.17} $\pm$2.03 & \textbf{74.40} $\pm$1.08 \\ 
H4 & 79.87 $\pm$1.34 & 78.15 $\pm$1.58 & 78.18 $\pm$0.98 & \underline{80.10} $\pm$1.00 & \textbf{84.05} $\pm$0.82 \\ 
H4ac & 55.22 $\pm$2.20 & 54.15 $\pm$2.96 & 51.80 $\pm$0.10 & \underline{59.26} $\pm$3.67 & \textbf{67.20} $\pm$1.05 \\ 
\hline

\textbf{\textit{Regulatory}} & & & & & \\ \hline
Enhancer & 54.51 $\pm$1.94 & 53.13 $\pm$4.52 & 52.50 $\pm$1.44 & \underline{55.20} $\pm$2.56 & \textbf{57.95} $\pm$1.10 \\ 
Enhancer Types & 43.36 $\pm$1.75 & \underline{48.16} $\pm$2.48 & 44.32 $\pm$1.18 & 47.17 $\pm$2.85 & \textbf{48.20} $\pm$2.30 \\ 
Promoter All & \underline{96.82} $\pm$0.47 & 95.57 $\pm$0.18 & 96.23 $\pm$0.17 & 96.65 $\pm$0.16 & \textbf{97.56} $\pm$0.16 \\ 
Promoter non-TATA & \underline{97.45} $\pm$0.69 & 95.86 $\pm$0.37 & 97.17 $\pm$0.17 & 96.31 $\pm$0.50 & \textbf{97.79} $\pm$0.18 \\ 
Promoter TATA & \underline{96.53} $\pm$0.81 & 95.88 $\pm$0.53 & 96.99 $\pm$0.49 & 96.21 $\pm$0.81 & \textbf{97.16} $\pm$0.32 \\ 
\hline

\textbf{\textit{Splice sites}} & & & & & \\ \hline
Splice Acceptor & \underline{97.99} $\pm$0.66 & 96.98 $\pm$0.49 & 97.49 $\pm$0.36 & 94.21 $\pm$0.37 & \textbf{98.13} $\pm$0.35 \\ 
Splice Donor & \textbf{98.50} $\pm$0.43 & 95.27 $\pm$1.07 & 94.33 $\pm$0.27 & 94.69 $\pm$0.67 & \underline{97.95} $\pm$0.38 \\ 
Splice All & \textbf{98.15} $\pm$1.01 & 94.05 $\pm$1.08 & 93.75 $\pm$1.25 & 92.87 $\pm$1.73 & \underline{97.70} $\pm$0.26 \\ 
\hline 
\end{tabular}
}
\end{table}

Table~\ref{tab:nt_benchmark} presents the Nucleotide Transformer benchmark results. 
Wisteria achieves the best result on 16 of the 18 tasks, indicating strong overall performance on this benchmark. 
Notably, it outperforms both attention based and Mamba based baselines on all histone mark prediction tasks.
This result is meaningful because histone mark prediction is often strongly associated with local sequence motifs and nearby regulatory context, which can be captured from relatively short genomic regions~\cite{whitaker2015epigenome}.
The consistent gains on these tasks therefore suggest that Wisteria is effective not only for long range dependency modeling but also for local epigenomic pattern recognition.

\subsubsection{Variant Effect Prediction}
\label{sec:vep}
To assess whether Wisteria can capture the long range dependencies underlying gene regulation, we performed experiments on the variant effect prediction (VEP) task.
This experiment uses the single nucleotide polymorphism (SNP) effect prediction dataset presented by~\cite{kao2024advancing}.
Biological evidence suggests that this task involves long range regulatory interactions~\cite{furlong2018developmental}.
The input of this dataset are sequences that centered around SNP loci in the human reference genome. 
The labels are calculated by the SuSiE tool~\cite{wang2020simple} (samples with probabilities >0.9 are marked as positive) with causal probabilities.
We adopt the same dataset split method.
Following the Caduceus evaluation protocol~\cite{schiff2024caduceus}, for each distance bucket, we sample 5,000 training points and fit an SVM classifier with an RBF kernel to predict VEP annotations.
We report the test AUROC as mean $\pm$ standard deviation over classifiers fitted on five random training subsets. \rev{For this experiment, we use the 768-dim configuration pretrained on 131,072 bp sequences.}
AUROC (Area Under the Receiver Operating Characteristic Curve) measures the model's ability to distinguish between positive and negative samples across all classification thresholds, with values ranging from 0 to 1, where 1.0 indicates perfect discrimination and 0.5 corresponds to random guessing.

We compared Wisteria with four baseline models. 
For this long range task, \textbf{HyenaDNA} uses the medium configuration with support for 160k bp sequences, and \textbf{Caduceus-Ph} uses a larger configuration with 16 MambaDNA layers and a hidden dimension of 256, pretrained on 131,072 bp sequences. 
\textbf{NT-v2} uses the same 500M multi-species configuration as described in Sec.~\ref{sec:nucleotide}.
The \textbf{Enformer}~\cite{avsec2021effective} employs a hybrid CNN and Transformer architecture that processes 196{,}608 bp input sequences.

In this experiment, we follow the same methodology and extract embeddings by mean pooling a 1536 bp window centered at the SNP location for both the reference and alternative sequences, then concatenating the resulting representations along the channel dimension. Caduceus-Ph, HyenaDNA, and Wisteria use base-pair tokenization, for which the window consists of 1536 tokens. 
Because Nucleotide Transformer was trained with 6-mer tokenization, the corresponding window spans 256 bp. 
For Enformer model, the final embedding has a receptive field of 128 bp, so a window of 12 tokens/positions is used. 
Caduceus-Ph, HyenaDNA, and Wisteria take 131k bp inputs, Nucleotide Transformer uses 12k bp inputs, and Enformer processes 196k bp inputs. 
To consistent with Caduceus, we performed hyperparameter optimization for our model within each distance category. 
We considered three regularization strengths for the set ${1, 5, 10}$, and selected the value with the highest average AUROC over five random-seed runs. 
Table~\ref{tab:hyperparams1} lists the selected regularization strength for each model.
 
\begin{table}[h]
\centering
\caption{SVM regularization strengths used for the variant effect prediction task.}
\label{tab:hyperparams1}
\resizebox{0.5\linewidth}{!}{
\begin{tabular}{lcccc}
\toprule
\textbf{Model} & \textbf{0 - 30k} & \textbf{30 - 100k} & \textbf{100k+} \\
\midrule
Enformer & 1 & 1 & 5  \\  
NT-v2 & 1 & 1 & 10  \\             
HyenaDNA & 1 & 1 & 5   \\  
Caduceus-Ph & 1 & 5 & 10  \\
Wisteria & 1 & 1 & 10  \\
\bottomrule
\end{tabular}
}
\end{table}

\begin{table}[htbp]
\centering
\caption{\rev{\textbf{Variant effect prediction results.} AUROC (\(\uparrow\)) is reported for the 0--30k, 30--100k, and 100k+ distance-to-TSS bins relative to the nearest transcription start site (TSS). Best values per task are bolded, second-best values are underlined, and \(\pm\) indicates the standard deviation across five SVM classifiers.}}
\begin{tabular}{lccc}
\hline
\textbf{Model} & \textbf{0--30k} & \textbf{30--100k} & \textbf{100k+} \\
\hline
Caduceus-Ph & \underline{0.678} $\pm$ 0.002 & 0.648 $\pm$ 0.016 & \underline{0.580} $\pm$ 0.011 \\
Enformer    & 0.668 $\pm$ 0.002 & \underline{0.659} $\pm$ 0.016 & 0.563 $\pm$ 0.032 \\
HyenaDNA    & 0.670 $\pm$ 0.002 & 0.599 $\pm$ 0.016 & 0.565 $\pm$ 0.024 \\
NT-v2       & 0.671 $\pm$ 0.002 & 0.593 $\pm$ 0.022 & 0.540 $\pm$ 0.011 \\
Wisteria        & \textbf{0.681} $\pm$ 0.003 & \textbf{0.663} $\pm$ 0.010 & \textbf{0.604} $\pm$ 0.003 \\
\hline
\end{tabular}
\label{tab:sequence_length_comparison}
\end{table}

We compare Wisteria with HyenaDNA, Caduceus-Ph, Nucleotide Transformer, and the supervised baseline Enformer. 
Table~\ref{tab:sequence_length_comparison} shows that Wisteria consistently outperforms HyenaDNA and NT-v2. 
In Table~\ref{tab:sequence_length_comparison}, 
In the 100k+ distance-to-TSS bin, Wisteria also achieves the strongest AUROC among all compared models.

\subsubsection{BEND Benchmark}
\label{sec:bend}

We evaluate Wisteria on the BEND benchmark~\cite{marinbend}, which is designed to assess genomic language models on biologically meaningful tasks derived from the human genome.
BEND provides a standardized set of seven curated tasks spanning different length scales,
from nucleotide-level prediction (e.g., gene finding over 1,433--14,000 bp) to sequence-level classification (e.g., chromatin accessibility over 512 bp).
These tasks, including enhancer annotation, histone modification prediction, CpG methylation and noncoding variant effects,
evaluate models' ability to handle realistic challenges such as long range dependencies, signal sparsity, and multi-label outputs.
Collectively, these tasks assess a model's ability to capture gene structure, regulatory elements, epigenetic modifications, and variant impacts while integrating local and distant sequence signals.

The seven BEND tasks follow two evaluation protocols.
The first five tasks---gene finding, enhancer annotation, chromatin accessibility, histone modification, and CpG methylation---are evaluated under a \textit{frozen backbone} protocol, in which the pretrained model is kept fixed and a lightweight task specific prediction head is trained on top.
The remaining two tasks, VEP for expression quantitative trait loci and disease associated noncoding variants, are evaluated in a \textit{zero-shot} setting, where model outputs are computed directly from frozen representations without any task specific fine-tuning.
\rev{For this experiment, we use the 768-dim configuration pretrained on 131,072 bp sequences.}
\rev{Following the benchmark protocol, our frozen backbone evaluations on the first five BEND tasks used a learning rate of $1 \times 10^{-4}$ and a batch size of 64 for the task specific lightweight prediction head.}

We compared Wisteria with six baseline models.
The \textbf{DNABERT} baseline follows the BERT base configuration with 12 layers, a 768-dimensional hidden size, and 6-mer tokenization, pretrained on the human reference genome with a 512 bp input window.
The \textbf{GENA-LM}~\cite{fishman2023gena} baseline uses the BigBird configuration with 12 layers and a 768-dimensional hidden size.
The \textbf{DNABERT-2} and \textbf{NT-v2} baselines use the same configurations as described in Sec.~\ref{sec:nucleotide}.
The \textbf{HyenaDNA} baseline for BEND uses the large model configuration, supporting sequences up to 1{,}000{,}000 bp with single-nucleotide resolution.
The \textbf{Caduceus-Ph} configuration is the same as described in Section~\ref{sec:genomics}.

\begin{table}[htbp]
\centering
\caption{\rev{\textbf{BEND benchmark results.} MCC (\(\uparrow\)) is reported for gene finding, AUPRC (\(\uparrow\)) for enhancer annotation, and AUROC (\(\uparrow\)) for all other tasks. Best values per task are bolded and second-best values are underlined.}}
\label{tab:model_performance_transposed}
\resizebox{\textwidth}{!}{
\begin{tabular}{lcccccc}
\toprule
\textbf{Task} & \textbf{DNABERT} & \textbf{DNABERT-2} & \textbf{NT-v2} & \textbf{GENA-LM} & \textbf{HyenaDNA} & \textbf{Wisteria} \\
\midrule
Gene finding              & 0.20 & 0.43 & \underline{0.64} & 0.52 & 0.35 & \textbf{0.67} \\
Enhancer annotation       & 0.03 & 0.03 & \underline{0.05} & 0.03 & 0.03 & \textbf{0.05} \\
Chromatin accessibility   & \textbf{0.85} & 0.81 & 0.80 & 0.76 & \underline{0.84} & \textbf{0.85} \\
Histone modification      & 0.79 & 0.78 & 0.76 & 0.78 & 0.76 & \textbf{0.84} \\
CpG methylation           & 0.91 & 0.90 & 0.91 & 0.91 & 0.91 & \textbf{0.93} \\
Variant effects (expression) & \underline{0.60} & 0.49 & 0.48 & 0.49 & 0.51 & \textbf{0.62} \\
Variant effects (disease)    & 0.56 & \underline{0.51} & 0.48 & 0.55 & 0.45 & \textbf{0.75} \\
\bottomrule
\end{tabular}
}
\end{table}

Table~\ref{tab:model_performance_transposed} shows that Wisteria achieves the strongest or tied-strongest performance across the seven BEND tasks, with notable gains in gene finding, histone modification, methylation, and variant effect prediction.
AUPRC (Area Under the Precision-Recall Curve) is used for the enhancer annotation task; it measures the trade off between precision and recall across thresholds and is particularly informative for imbalanced datasets where positive samples are rare, with values closer to 1.0 indicating superior performance.
These results support the ability of Wisteria to model complex genomic dependencies across tasks involving both short and long range interactions, sparse signals, and hierarchical regulatory features.

\subsection{\rev{Ablation Study}}
\label{sec:ablation}

\rev{To validate the design choices of Wisteria, we conduct two ablation studies on the Nucleotide Transformer tasks. 
The first isolates the contribution of the core architectural modules, and the second examines the allocation of shallow GCMB blocks used in the final model.}

\subsubsection{\rev{Component Ablation on Nucleotide Transformer tasks}}
\label{sec:nt_ablation}

\rev{To assess the contribution of the core components, we conducted a component ablation study on the Nucleotide Transformer tasks. 
The Nucleotide Transformer benchmark covers a relatively diverse set of downstream tasks, making it suitable for evaluating the task level generalization of each architectural component. 
Starting from the complete \textit{Wisteria} architecture, we remove individual modules to obtain three reduced variants: \textit{w/o Fourier}, which removes the Fourier based attention layer; \textit{w/o GCMB}, which replaces the shallow GCMB blocks in the first five layers with standard BiMamba blocks; and \textit{w/o GCMB \& Gated MLP}, which removes both the GCMB blocks and the gated MLP refinement, leaving only the BiMamba backbone. 
The metrics follow the Nucleotide Transformer benchmark protocol. }
\rev{Table~\ref{tab:nt_ablation} summarizes the results for each task in this ablation study on the Nucleotide Transformer tasks.}

\begin{table}[htbp]
\centering
\caption{\rev{\textbf{Ablation study results on Nucleotide Transformer tasks.} Best values are highlighted in bold, and second-best values are underlined.}}
\label{tab:nt_ablation}
\resizebox{\textwidth}{!}{
\begin{tabular}{l c c c c}
\hline
 & \textcolor{blue}{\textbf{Wisteria (Full)}} & \textcolor{blue}{\textbf{w/o Fourier}} & \textcolor{blue}{\textbf{w/o GCMB}} & \textcolor{blue}{\textbf{w/o GCMB \& Gated MLP}} \\
\hline
\textcolor{blue}{\textbf{\textit{Histone}}} & & & & \\ \hline
\textcolor{blue}{H3} & \textcolor{blue}{\textbf{84.47} $\pm$0.72} & \textcolor{blue}{\underline{83.81} $\pm$0.71} & \textcolor{blue}{81.24 $\pm$1.62} & \textcolor{blue}{78.21 $\pm$2.50} \\
\textcolor{blue}{H3K4me1} & \textcolor{blue}{\textbf{59.39} $\pm$0.89} & \textcolor{blue}{\underline{57.94} $\pm$4.63} & \textcolor{blue}{55.11 $\pm$2.55} & \textcolor{blue}{52.44 $\pm$1.13} \\
\textcolor{blue}{H3K4me2} & \textcolor{blue}{\textbf{55.36} $\pm$1.83} & \textcolor{blue}{\underline{53.96} $\pm$2.92} & \textcolor{blue}{45.97 $\pm$2.73} & \textcolor{blue}{49.38 $\pm$2.22} \\
\textcolor{blue}{H3K4me3} & \textcolor{blue}{\textbf{64.27} $\pm$0.85} & \textcolor{blue}{\underline{62.99} $\pm$3.31} & \textcolor{blue}{57.39 $\pm$3.16} & \textcolor{blue}{57.46 $\pm$1.87} \\
\textcolor{blue}{H3K9ac} & \textcolor{blue}{\textbf{67.65} $\pm$1.55} & \textcolor{blue}{\underline{66.14} $\pm$1.37} & \textcolor{blue}{64.49 $\pm$1.79} & \textcolor{blue}{63.14 $\pm$1.44} \\
\textcolor{blue}{H3K14ac} & \textcolor{blue}{\textbf{70.16} $\pm$1.90} & \textcolor{blue}{\underline{68.40} $\pm$4.91} & \textcolor{blue}{64.81 $\pm$2.33} & \textcolor{blue}{61.54 $\pm$2.15} \\
\textcolor{blue}{H3K36me3} & \textcolor{blue}{\textbf{71.55} $\pm$1.02} & \textcolor{blue}{\underline{69.57} $\pm$2.32} & \textcolor{blue}{63.64 $\pm$1.14} & \textcolor{blue}{62.62 $\pm$1.79} \\
\textcolor{blue}{H3K79me3} & \textcolor{blue}{\textbf{74.40} $\pm$1.08} & \textcolor{blue}{\underline{73.68} $\pm$0.42} & \textcolor{blue}{73.13 $\pm$3.66} & \textcolor{blue}{68.37 $\pm$2.32} \\
\textcolor{blue}{H4} & \textcolor{blue}{\textbf{84.05} $\pm$0.82} & \textcolor{blue}{\underline{81.92} $\pm$1.01} & \textcolor{blue}{80.57 $\pm$1.58} & \textcolor{blue}{81.27 $\pm$1.33} \\
\textcolor{blue}{H4ac} & \textcolor{blue}{\textbf{67.20} $\pm$1.05} & \textcolor{blue}{\underline{65.23} $\pm$0.13} & \textcolor{blue}{63.51 $\pm$2.91} & \textcolor{blue}{61.38 $\pm$2.15} \\
\hline
\textcolor{blue}{\textbf{\textit{Regulatory}}} & & & & \\ \hline
\textcolor{blue}{Enhancer} & \textcolor{blue}{\textbf{57.95} $\pm$1.10} & \textcolor{blue}{\underline{54.20} $\pm$1.46} & \textcolor{blue}{54.13 $\pm$4.51} & \textcolor{blue}{53.25 $\pm$1.94} \\
\textcolor{blue}{Enhancer Types} & \textcolor{blue}{\textbf{48.20} $\pm$2.30} & \textcolor{blue}{\underline{45.78} $\pm$1.28} & \textcolor{blue}{45.32 $\pm$2.42} & \textcolor{blue}{43.26 $\pm$1.77} \\
\textcolor{blue}{Promoter All} & \textcolor{blue}{\textbf{97.56} $\pm$0.16} & \textcolor{blue}{\underline{97.15} $\pm$0.13} & \textcolor{blue}{95.74 $\pm$0.25} & \textcolor{blue}{95.79 $\pm$0.42} \\
\textcolor{blue}{Promoter non-TATA} & \textcolor{blue}{\textbf{97.79} $\pm$0.18} & \textcolor{blue}{\underline{97.26} $\pm$0.12} & \textcolor{blue}{96.46 $\pm$0.36} & \textcolor{blue}{96.44 $\pm$0.64} \\
\textcolor{blue}{Promoter TATA} & \textcolor{blue}{\textbf{97.16} $\pm$0.32} & \textcolor{blue}{\underline{96.80} $\pm$0.49} & \textcolor{blue}{95.72 $\pm$0.51} & \textcolor{blue}{95.97 $\pm$0.83} \\
\hline
\textcolor{blue}{\textbf{\textit{Splice sites}}} & & & & \\ \hline
\textcolor{blue}{Splice Acceptor} & \textcolor{blue}{\textbf{98.13} $\pm$0.35} & \textcolor{blue}{\underline{97.51} $\pm$0.31} & \textcolor{blue}{96.32 $\pm$0.53} & \textcolor{blue}{96.67 $\pm$0.67} \\
\textcolor{blue}{Splice Donor} & \textcolor{blue}{\textbf{97.95} $\pm$0.38} & \textcolor{blue}{\underline{97.82} $\pm$0.22} & \textcolor{blue}{95.19 $\pm$1.06} & \textcolor{blue}{94.16 $\pm$0.45} \\
\textcolor{blue}{Splice All} & \textcolor{blue}{\textbf{97.70} $\pm$0.26} & \textcolor{blue}{\underline{96.98} $\pm$1.28} & \textcolor{blue}{93.45 $\pm$1.08} & \textcolor{blue}{92.32 $\pm$1.09} \\
\hline
\end{tabular}
}
\end{table}

\rev{As shown in Table~\ref{tab:nt_ablation}, removing the Fourier based attention layer (w/o Fourier) leads to consistent but moderate drops across most tasks, indicating that global contextual integration contributes meaningfully to prediction quality. 
Further replacing the shallow GCMB blocks with standard BiMamba blocks (w/o GCMB) causes a larger decline, particularly on histone mark and splice site prediction, confirming that early stage multi scale feature extraction is a key driver of performance. 
The most reduced variant (w/o GCMB \& Gated MLP) exhibits the largest degradation, demonstrating that the combination of GCMB and gated MLP refinement provides substantial gains over the BiMamba backbone alone. 
Because the magnitude of the drop varies across tasks, we conservatively interpret these results as evidence that the three components provide complementary modeling biases rather than as proof that any single module is universally dominant. 
Broader downstream architectural validation across additional settings remains a direction for future work.}

\subsubsection{\rev{N-block Allocation on Nucleotide Transformer tasks}}
\label{sec:nblock_ablation}

\rev{To examine the block allocation in Figure~\ref{fig:framework}, we conducted an ablation study on the Nucleotide Transformer tasks over the number of shallow GCMB blocks. We compared $N{=}0$, $N{=}2$, $N{=}5$, and $N{=}8$ while keeping the remaining training and evaluation protocol unchanged. Here, $N{=}5$ corresponds to the final Wisteria configuration used in Figure~\ref{fig:framework}. 
Table~\ref{tab:nt_nblock_full} reports the results for all tasks on the Nucleotide Transformer benchmark, covering histone mark prediction, regulatory classification, promoter identification, and splice-site prediction.}

\begin{table}[htbp]
\centering
\caption{\rev{\textbf{N-block Allocation results on Nucleotide Transformer tasks.} $N$ denotes the number of shallow GCMB blocks. The $N{=}5$ configuration corresponds to the final Wisteria architecture used in Figure~\ref{fig:framework}. Best values are highlighted in bold, and second-best values are underlined.}}
\label{tab:nt_nblock_full}
\resizebox{\textwidth}{!}{
\begin{tabular}{l c c c c}
\hline
 & \textcolor{blue}{\textbf{$N{=}0$}} & \textcolor{blue}{\textbf{$N{=}2$}} & \textcolor{blue}{\textbf{$N{=}5$}} & \textcolor{blue}{\textbf{$N{=}8$}} \\
\hline
\textcolor{blue}{\textbf{\textit{Histone}}} & & & & \\ \hline
\textcolor{blue}{H3} & \textcolor{blue}{81.58 $\pm$1.12} & \textcolor{blue}{\underline{83.06} $\pm$0.75} & \textcolor{blue}{\textbf{84.47} $\pm$0.72} & \textcolor{blue}{84.31 $\pm$0.71} \\
\textcolor{blue}{H3K4me1} & \textcolor{blue}{55.32 $\pm$2.05} & \textcolor{blue}{\underline{57.86} $\pm$1.34} & \textcolor{blue}{\textbf{59.39} $\pm$0.89} & \textcolor{blue}{59.18 $\pm$0.96} \\
\textcolor{blue}{H3K4me2} & \textcolor{blue}{46.54 $\pm$2.90} & \textcolor{blue}{\underline{50.88} $\pm$2.41} & \textcolor{blue}{\textbf{55.36} $\pm$1.83} & \textcolor{blue}{55.14 $\pm$1.85} \\
\textcolor{blue}{H3K4me3} & \textcolor{blue}{57.88 $\pm$2.18} & \textcolor{blue}{\underline{61.92} $\pm$1.45} & \textcolor{blue}{\textbf{64.27} $\pm$0.85} & \textcolor{blue}{64.09 $\pm$0.92} \\
\textcolor{blue}{H3K9ac} & \textcolor{blue}{64.76 $\pm$1.89} & \textcolor{blue}{66.38 $\pm$1.52} & \textcolor{blue}{\underline{67.65} $\pm$1.55} & \textcolor{blue}{\textbf{67.72} $\pm$1.44} \\
\textcolor{blue}{H3K14ac} & \textcolor{blue}{65.18 $\pm$2.56} & \textcolor{blue}{\underline{68.46} $\pm$2.10} & \textcolor{blue}{\textbf{70.16} $\pm$1.90} & \textcolor{blue}{69.94 $\pm$1.91} \\
\textcolor{blue}{H3K36me3} & \textcolor{blue}{64.02 $\pm$2.04} & \textcolor{blue}{67.12 $\pm$1.68} & \textcolor{blue}{\underline{71.55} $\pm$1.02} & \textcolor{blue}{\textbf{71.63} $\pm$0.95} \\
\textcolor{blue}{H3K79me3} & \textcolor{blue}{73.42 $\pm$1.38} & \textcolor{blue}{\underline{73.96} $\pm$0.96} & \textcolor{blue}{\textbf{74.40} $\pm$1.08} & \textcolor{blue}{74.26 $\pm$1.06} \\
\textcolor{blue}{H4} & \textcolor{blue}{80.71 $\pm$1.25} & \textcolor{blue}{82.88 $\pm$0.88} & \textcolor{blue}{\underline{84.05} $\pm$0.82} & \textcolor{blue}{\textbf{84.12} $\pm$0.72} \\
\textcolor{blue}{H4ac} & \textcolor{blue}{63.69 $\pm$1.76} & \textcolor{blue}{\underline{65.84} $\pm$1.18} & \textcolor{blue}{\textbf{67.20} $\pm$1.05} & \textcolor{blue}{67.06 $\pm$1.08} \\
\hline
\textcolor{blue}{\textbf{\textit{Regulatory}}} & & & & \\ \hline
\textcolor{blue}{Enhancer} & \textcolor{blue}{54.46 $\pm$1.15} & \textcolor{blue}{\underline{54.92} $\pm$1.08} & \textcolor{blue}{\textbf{57.95} $\pm$1.10} & \textcolor{blue}{57.84 $\pm$0.92} \\
\textcolor{blue}{Enhancer Types} & \textcolor{blue}{45.51 $\pm$1.62} & \textcolor{blue}{\underline{46.92} $\pm$1.35} & \textcolor{blue}{\underline{48.20} $\pm$2.30} & \textcolor{blue}{\textbf{48.27} $\pm$1.12} \\
\textcolor{blue}{Promoter All} & \textcolor{blue}{96.08 $\pm$0.22} & \textcolor{blue}{\underline{97.02} $\pm$0.15} & \textcolor{blue}{\textbf{97.56} $\pm$0.16} & \textcolor{blue}{97.49 $\pm$0.13} \\
\textcolor{blue}{Promoter non-TATA} & \textcolor{blue}{96.71 $\pm$0.19} & \textcolor{blue}{97.34 $\pm$0.14} & \textcolor{blue}{\underline{97.79} $\pm$0.18} & \textcolor{blue}{\textbf{97.83} $\pm$0.10} \\
\textcolor{blue}{Promoter TATA} & \textcolor{blue}{96.08 $\pm$0.28} & \textcolor{blue}{\underline{96.78} $\pm$0.18} & \textcolor{blue}{\textbf{97.16} $\pm$0.32} & \textcolor{blue}{97.11 $\pm$0.16} \\
\hline
\textcolor{blue}{\textbf{\textit{Splice sites}}} & & & & \\ \hline
\textcolor{blue}{Splice Acceptor} & \textcolor{blue}{96.74 $\pm$0.32} & \textcolor{blue}{\underline{97.34} $\pm$0.28} & \textcolor{blue}{\textbf{98.13} $\pm$0.35} & \textcolor{blue}{98.02 $\pm$0.24} \\
\textcolor{blue}{Splice Donor} & \textcolor{blue}{95.47 $\pm$0.42} & \textcolor{blue}{96.86 $\pm$0.35} & \textcolor{blue}{\underline{97.95} $\pm$0.38} & \textcolor{blue}{\textbf{98.01} $\pm$0.25} \\
\textcolor{blue}{Splice All} & \textcolor{blue}{93.82 $\pm$0.58} & \textcolor{blue}{\underline{95.12} $\pm$0.45} & \textcolor{blue}{\textbf{97.70} $\pm$0.26} & \textcolor{blue}{97.56 $\pm$0.32} \\
\hline
\end{tabular}
}
\end{table}

\rev{The results in Table~\ref{tab:nt_nblock_full} show a consistent trend across the full Nucleotide Transformer tasks. 
Compared with $N{=}0$ and $N{=}2$, the $N{=}5$ configuration yields clear gains on most tasks. 
Increasing the GCMB allocation further to $N{=}8$ produces only marginal changes relative to $N{=}5$, with several tasks showing near-tied or slightly higher values for $N{=}8$.
This pattern suggests that most of the benefit is already captured by the $N{=}5$ setting. 
Consequently, the first five GCMB layers serve as a shallow local modeling stage. With kernel size 9 and progressively expanded dilation rates, they cover local to mid range patterns ranging from short motifs to short regulatory elements. The deeper layers remain available for broader contextual modeling by BiMamba, gated MLP refinement, and the final Fourier based attention module. Together, the empirical trend and this design rationale support the use of five GCMB layers as a practical trade off between local feature extraction and broader contextual modeling.}

\subsection{Analysis}
\label{sec:analysis}

Beyond benchmark results, we further analyze three design related aspects of Wisteria: positional encoding, local feature modeling, and computational efficiency.

\subsubsection{Comparison of Fourier and Rotary Position Embeddings}
\label{sec:fope_vs_rope}

To further investigate the impact of positional encoding on length generalization, we conducted a comparison between Fourier Position Embedding (FoPE) and Rotary Position Embedding (RoPE). 
Both models were pretrained with a maximum sequence length of 1024 on the hg38 dataset using the Wisteria-100M architecture, and subsequently evaluated across varying input lengths.

\begin{figure}[H]
    \centering
    \includegraphics[width=0.7\textwidth]{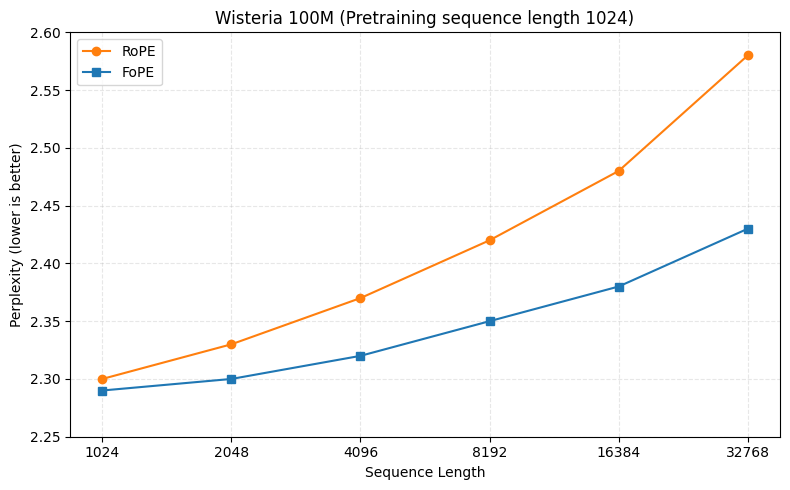}
    \caption{Comparison of Fourier Position Embedding (FoPE) and Rotary Position Embedding (RoPE) in pretraining on the hg38 dataset using Wisteria-100M with different length. }
    \label{fig:fope_vs_rope}
\end{figure}

As shown in Figure~\ref{fig:fope_vs_rope}, FoPE consistently achieves lower perplexity than RoPE across all tested sequence lengths, with the advantage becoming more pronounced for longer inputs ($> 8$k).
This result demonstrates that FoPE possesses stronger length generalization capability and superior modeling of long range dependencies.
The performance gain further amplifies as the receptive field expands, indicating that FoPE provides more stable and expressive representations for extended contextual information.

\subsubsection{Evaluation of the \rev{Gated Convolution Branch} for Local Feature Modeling on Short Sequences}
\label{sec:cnn_local}

\begin{figure}[H]
    \centering
    \begin{subfigure}[b]{0.45\textwidth}
        \centering
        \includegraphics[width=\textwidth]{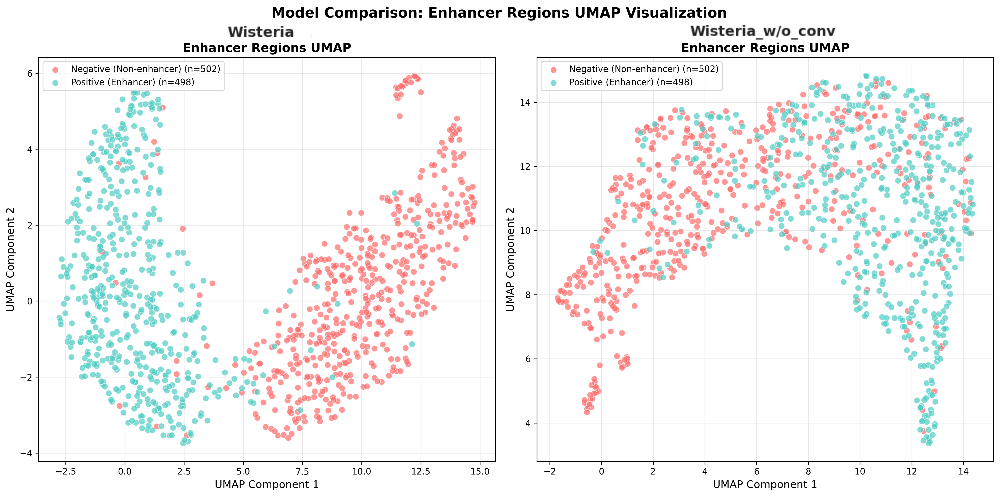}
        \caption{Enhancer}
        \label{fig:enhancer}
    \end{subfigure}
    \hfill
    \begin{subfigure}[b]{0.45\textwidth}
        \centering
        \includegraphics[width=\textwidth]{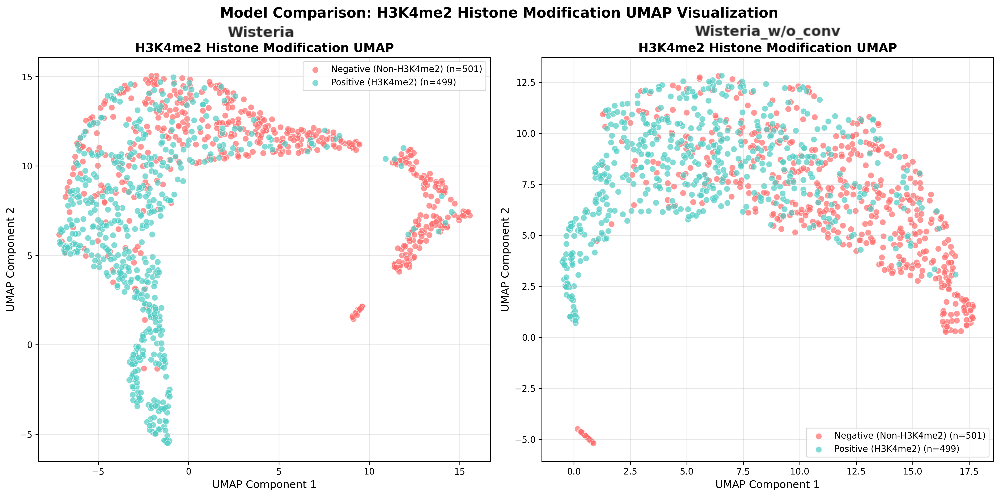}
        \caption{H3K4me2 Histone}
        \label{fig:h3k4me2}
    \end{subfigure}

    \vspace{1cm} 

    \begin{subfigure}[b]{0.45\textwidth}
        \centering
        \includegraphics[width=\textwidth]{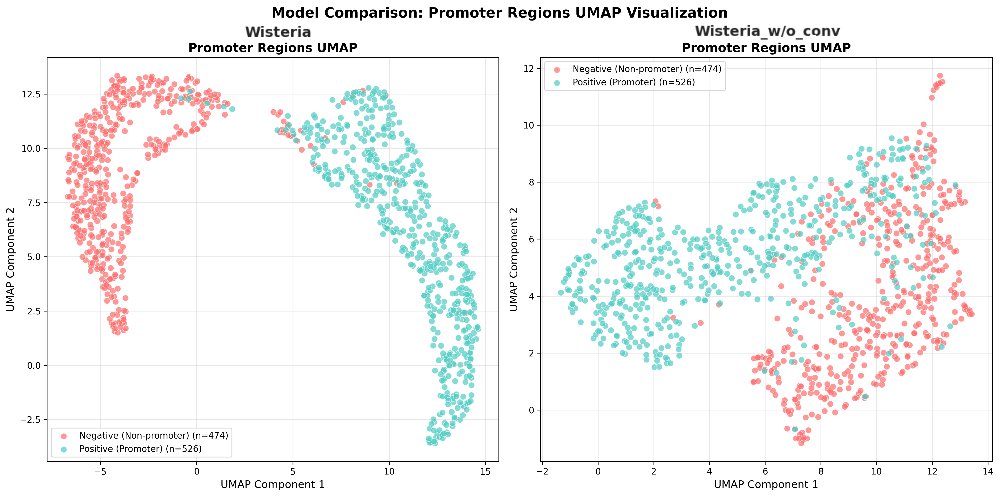}
        \caption{Promoter}
        \label{fig:promoter}
    \end{subfigure}
    \hfill
    \begin{subfigure}[b]{0.45\textwidth}
        \centering
        \includegraphics[width=\textwidth]{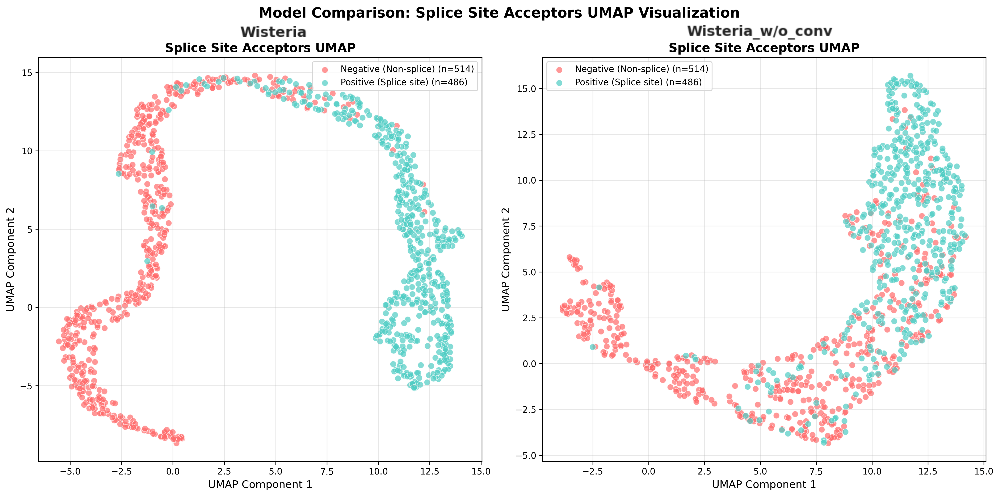}
        \caption{Splice Site}
        \label{fig:splice_site}
    \end{subfigure}

    \caption{\rev{UMAP visualization of embeddings for short sequence regions with and without the gated convolution branch.}}
    \label{fig:umap_visualizations}
\end{figure}

To assess the efficacy of the \rev{gated convolution branch within the GCMB module} for modeling local features of short genomic sequences, we conducted four comparative experiments using enhancer, H3K4me2 histone, promoter, and splice-site datasets. 
\rev{The four subplots in Figure~\ref{fig:umap_visualizations} were all generated using the same dual model comparison setup, where embeddings from models with and without the gated convolution branch were extracted for 1000 balanced samples per dataset. 
For all four subplots, we used mean pooling over sequence embeddings, standardized the pooled representations with StandardScaler, and applied two-dimensional Uniform Manifold Approximation and Projection (UMAP) with $n\_neighbors=15$, $\textit{min\_dist}=0.5$, Euclidean distance, and random seed 42.
The maximum sequence length was 1024 bp and the embedding extraction batch size was 8.} 
Figure~\ref{fig:umap_visualizations} shows that \rev{adding the gated convolution branch substantially improves} the separability of the four short sequence regions, like enhancers, H3K4me2 histone marks, promoters, and splice sites, indicating \rev{stronger local feature extraction.}

\subsubsection{Throughput and Peak Memory}
\label{sec:throughput}

\begin{figure}[H]
    \centering
    \begin{subfigure}{0.7\textwidth}
        \centering
        \includegraphics[width=\textwidth]{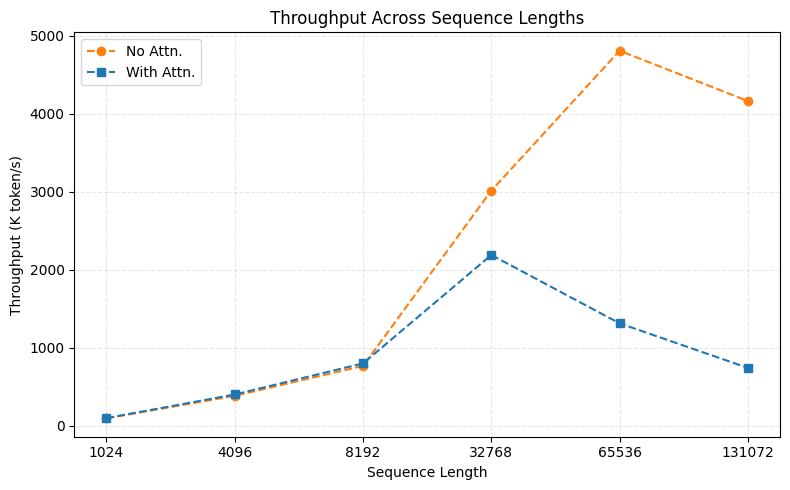}
        \caption{Throughput}
        \label{fig:sub1}
    \end{subfigure}
    \vspace{1em}
    \begin{subfigure}{0.7\textwidth}
        \centering
        \includegraphics[width=\textwidth]{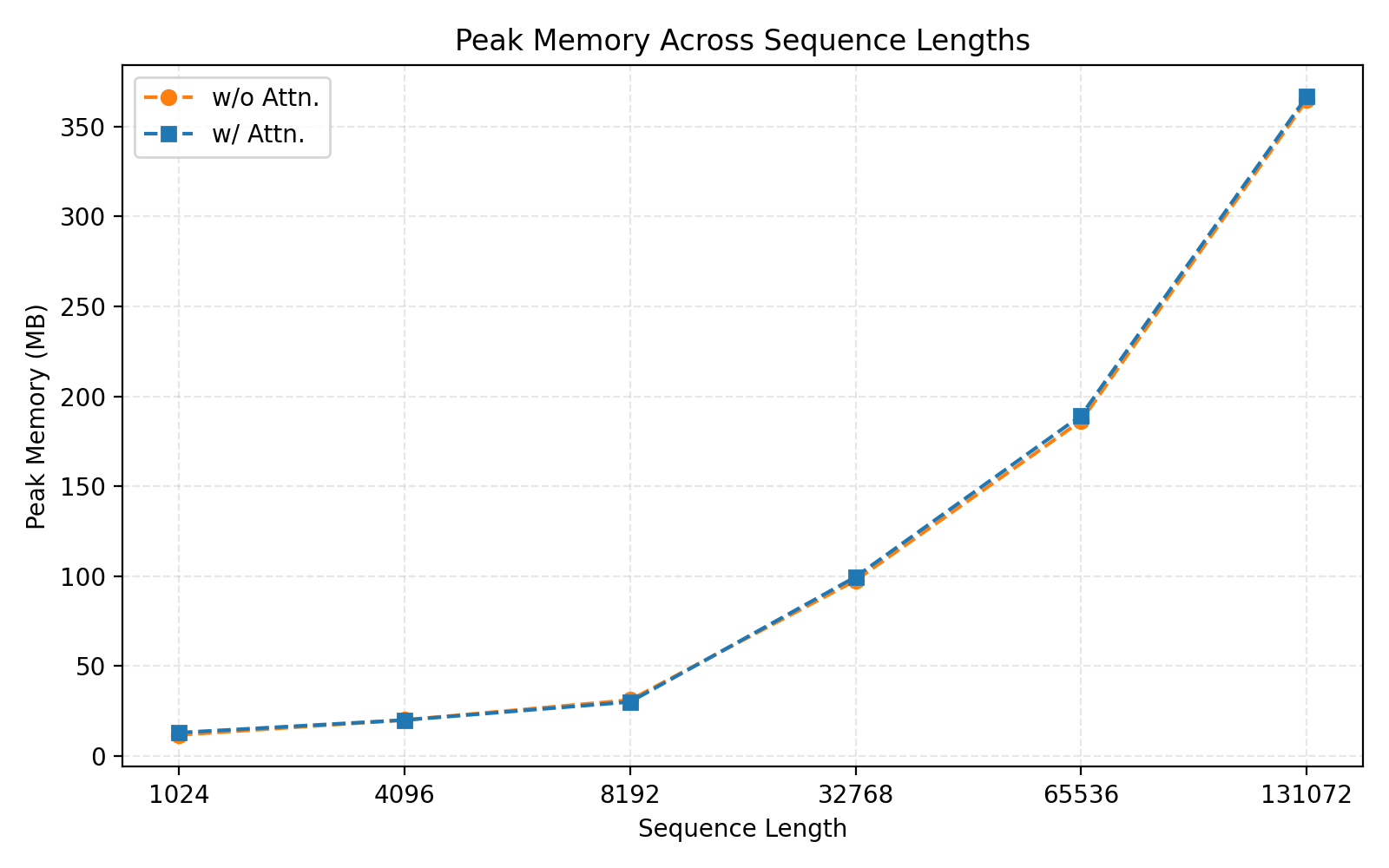}
        \caption{Peak Memory}
        \label{fig:sub2}
    \end{subfigure}
    \caption{Performance metrics at different sequence lengths. The line plots show the variation in throughput (K tokens/s) and peak memory (MB) with respect to sequence length for architectures with and without the attention layer.}
    \label{fig:twofigs}
\end{figure}

We conducted a quantitative analysis to evaluate the impact of the attention layer on the efficiency of modeling long DNA sequences. Figure~\ref{fig:twofigs} compares the full model with a variant in which the attention layer is replaced by bidirectional Mamba layers.

The results indicate that, with a FlashAttention-2 implementation~\cite{dao2023flashattention2}, the attention layer does not introduce additional peak memory in this setting. 
In terms of throughput, the two models exhibit comparable performance for sequence lengths ranging from 1k to 8k, which may be attributed to the GPU not being fully saturated, with bottlenecks stemming from data transfer or kernel launch latencies. 
When sequence lengths exceed 32k, the attention layer shows a decrease in computational efficiency due to its quadratic complexity. 
Nevertheless, because the attention layer improves downstream performance and the additional overhead remains within an acceptable range, we regard this trade off as practical for long sequence modeling.

\section{Conclusion}
\label{conclusion}

In this study, we presented a genomic language model based on multi scale feature learning for DNA sequences, named Wisteria.
Wisteria captures both local and global dependencies in DNA sequences through the combination of gated convolutions, selective state space model and Fourier based attention.
By integrating FoPE, Wisteria provides smooth frequency domain representations and strong length generalization, while gated MLPs and dilated convolutions support fine-grained local motif recognition.
\rev{Experiments on Genomic Benchmarks, Nucleotide Transformer tasks, BEND and variant effect prediction show that our model achieved strong overall performance. 
These results indicate that Wisteria is effective for regulatory annotation, splice-site prediction, zero-shot evaluation and long-context prediction tasks.}

\rev{Current limitations include the human centered training and evaluation setting and the exclusive use of the unsupervised MLM objective. 
Because pretraining is mainly conducted on the human reference genome hg38, the model may not fully capture evolutionary constraints and function related motifs that are better revealed through cross species conservation. In addition, the current study does not examine the benefits of supervised training signals. 
For example, Enformer~\cite{avsec2021effective} suggests that such signals may further improve genomic prediction. 
Future work will extend pretraining to additional species and investigate supervised objectives for stronger downstream performance.}

\subsubsection*{Acknowledgements}
This work is supported by Inner Mongolia Natural Science Foundation
(No.2024 M\newline S06013); 
Inner Mongolia Autonomous Region Science and Technology Programme Project (Nos.2025YFDZ0017, 2025KYPT0041, 2025KYPT0064);
the special research project of First-class Discipline of Inner Mongolia of China (YLXKZX-ND-036).

\bibliography{ref}

\end{document}